\documentclass[preprint,12pt]{elsarticle}

\usepackage{amssymb}
\usepackage{amsmath}
\usepackage{booktabs}
\usepackage{multirow}
\usepackage{amsmath}
\usepackage[hyphens]{url}
\usepackage{breakurl}
\usepackage{xurl}
\usepackage{url} 
\usepackage{hyperref}

\journal{Fertility and Sterility}

\begin{document}

\begin{frontmatter}



 \author{Zahra Asghari Varzaneh\corref{cor1}\fnref{label1}}
 \ead{zahra.asghari-varzaneh@mau.se}
\cortext[cor1]{Corresponding author}

\title{Predicting Blastocyst Formation in IVF: Integrating DINOv2 and Attention-Based LSTM on Time-Lapse Embryo Images} 


\author{Niclas Wölner-Hanssen\fnref{label2}}
\author{Reza Khoshkangini\fnref{label1}}
\author{Thomas Ebner\fnref{label3}}
\author{Magnus Johnsson\fnref{label4}}

\affiliation[label1]{
    organization={Sustainable Digitalisation Research Center, Department of Computer Science and Media Technology}, 
    addressline={Malmö University},
    city={Malmö},
    country={Sweden}
}

\affiliation[label2]{
    organization={School of Information Technology},
    addressline={Halmstad University},
    city={Halmstad},
    country={Sweden}
}

\affiliation[label3]{
    organization={Kepler Universitätsklinikum},
    city={Linz},
    country={Austria}
}

\affiliation[label4]{
    organization={Department of Computer Science},
    addressline={Kristianstad University},
    city={Kristianstad},
    country={Sweden}
}

\begin{abstract}
The selection of the optimal embryo for transfer is a critical yet challenging step in in vitro fertilization (IVF), primarily due to its reliance on the manual inspection of extensive time-lapse imaging data. A key obstacle in this process is predicting blastocyst formation from the limited number of daily images available. Many clinics also lack complete time-lapse systems, so full videos are often unavailable. In this study, we aimed to predict which embryos will develop into blastocysts using limited daily images from time-lapse recordings. We propose a novel hybrid model that combines DINOv2, a transformer-based vision model, with an enhanced long short-term memory (LSTM) network featuring a multi-head attention layer. DINOv2 extracts meaningful features from embryo images, and the LSTM model then uses these features to analyze embryo development over time and generate final predictions. We tested our model on a real dataset of 704 embryo videos. The model achieved 96.4\% accuracy, surpassing existing methods. It also performs well with missing frames, making it valuable for many IVF laboratories with limited imaging systems. Our approach can assist embryologists in selecting better embryos more efficiently and with greater confidence.\\
\end{abstract}






\end{frontmatter}



\section{Introduction}
\label{sec:introduction}

IVF is among the most widely used and effective treatments for infertility and for preventing transmission of genetic disorders \cite{eugster1999psychological}. The procedure involves retrieving mature oocytes, fertilizing them with sperm under controlled laboratory conditions, and transferring the resulting embryos to the uterus. A full IVF cycle spans roughly two to three weeks, though dividing the process into discrete stages can lengthen the overall timeline \cite{blake2005cleavage}.
Because many embryos fail to survive in vitro, clinicians routinely culture several embryos in parallel. Modern time-lapse incubators create a stable culture environment while continuously recording high-resolution images that capture each embryo’s morphologic progression \cite{lukassen2005two}. Most embryos reach the blastocyst stage in about five days, though some require up to seven; others arrest before blastocyst formation. Only embryos that successfully form blastocysts are considered for transfer or cryopreservation, making accurate, timely identification of this stage a pivotal decision point in IVF \cite{swain2014decisions}.

With the development of time-lapse incubators, it is possible to continuously monitor embryo development. This technology helps doctors study the shape and development of embryos by looking at images taken over time \cite{wong2013time,liao2021development}. These images are saved as a series of frames in a time-lapse video that shows how the embryo grows from zygote to blastocyst stage. Embryologists use these videos to annotate the dynamic development of preimplantation embryos and to additionally assess their morphogenetic features such as cell number, cell shape, symmetry, presence of fragments, and the appearance of the blastocyst \cite{machtinger2013morphological}. This manual evaluation helps them select the best embryo for transfer. However, this process is time-consuming, difficult, and expensive \cite{motato2016morphokinetic}. Therefore, artificial intelligence (AI) tools and algorithms are becoming necessary in fertility clinics. Recently, AI methods, especially deep learning models, have made great progress in understanding and learning from large-scale data \cite{varzaneh2022new,jamali2025context,varzaneh2025ensemble}. Deep learning achievements at the level of human experts or even better, have been reported in screening and diagnosing diseases with medical images \cite{shen2017deep,razzak2017deep}. Today, the integration of AI into IVF and embryo evaluation has significantly improved the accuracy and efficiency of these processes \cite{fernandez2020artificial}. 
With the help of AI, embryologists can better select which embryos have the best chance of implanting and resulting in a pregnancy. This helps reduce the number of IVF cycles needed and lowers both emotional and financial stress for patients. Several studies have used deep learning algorithms to support embryo evaluation \cite{luong2024artificial}.

However, most of them only focus on images from the final stage, named blastocyst formation, while the earlier and middle stages of embryo development are also very important. In this study, we aim to use time-lapse monitoring (TLM) images collected throughout embryo growth to predict blastocyst formation more accurately.
A key challenge is that many IVF laboratories still do not have full access to TLM systems. In some cases, image sequences may be incomplete due to interruptions in imaging \cite{swain2014decisions}. Therefore, there is a need for a model that can learn from embryo growth over multiple days, even when only a limited number of images are available.
Given the challenges mentioned above, we developed a novel hybrid model for predicting blastocyst formation by applying temporal patterns from sparse, daily embryo images as a common constraint in many clinical settings. Previous approaches often relied on complete TLM videos, which are not always available. Our method addresses this issue by introducing a framework that combines the representational power of a transformer-based feature extractor (DINOv2) with the sequential modeling capability of an LSTM equipped with multi-head attention. This model allows for effective learning from limited image sequences, reducing dependency on continuous imaging systems. The following research questions (RQs) further elaborate the investigative objectives of our proposed approach in this study:

•	\textbf{RQ1—Feature Extraction and Temporal Modeling:} To what extent can a hybrid architecture combining DINOv2 for spatial feature extraction and a multi-head attention LSTM for temporal modeling accurately predict blastocyst formation from a sequence of daily embryo images?

•	\textbf{RQ2—Attention Mechanism Contribution:} How does the integration of a multi-head attention mechanism into the LSTM network enhance the model's ability to focus on critical developmental stages and improve prediction accuracy over a standard sequential model?

Taken together, these research objectives aim to counter the limitations of existing methods by mapping sparse, daily morphological snapshots to a key developmental outcome. First, to answer RQ1, we develop and evaluate our hybrid DINOv2-LSTM-attention model, demonstrating its superior performance in capturing both spatial and temporal dependencies. Second, RQ2, provides an analytical insight into the inner workings of the model, using the attention weights to interpret which time points are most influential for the prediction, thereby validating the design choice of the multi-head attention layer. The development of this accessible and efficient predictive model for embryo selection constitutes this paper's main contribution.

The main contributions of this study are summarized as follows:
\begin{itemize}
    \item Developing DINOv2 to extract detailed morphological features from individual embryo images, preserving critical spatial and contextual information.
    \item Introducing a Temporal LSTM enhanced with multi-head attention to accurately model sequential growth patterns while maintaining computational efficiency.
    \item Reducing dependency on continuous imaging systems to evaluate high-quality embryos for a broader range of IVF laboratories.
    \item Designing a model to address incomplete or limited TLM data to process sparse input frames (one image per day) without requiring complex hardware equipment to handle large volumes of input frames.
\end{itemize}

This proposed hybrid model helps solve a practical problem in many IVF labs. Not all clinics have access to full-time-lapse imaging systems, and in some cases, parts of the video may be missing. Our model can still make accurate predictions even when only a few daily images are available. This reduces the need for complete time-lapse videos, lowers computational cost, and makes the method more accessible in real clinical environments.

\section{Related work}
IVF has transformed reproductive medicine; however, improving methods to select embryos for transfer is essential to increase success rates. Recent advances in AI, particularly in deep learning and machine learning, have provided innovative approaches for analyzing patient clinical data and time-lapse imaging of embryonic development. Researchers are increasingly adopting sophisticated models for automated embryo grading, novel architectures for analyzing temporal patterns in morphokinetics, and hybrid models that integrate multiple data sources to predict implantation potential. These techniques offer a morphological assessment that is faster and more accurate than traditional manual methods. Despite these advances, challenges and gaps in the literature still need to be addressed.

Liao et al. \cite{liao2021development} developed two ensemble models, named STEM and STEM+, to predict the potential of blastocysts using time-lapse videos. They integrated DenseNet201, LSTM networks, and gradient boosting classifiers to analyze the first three days of embryo development. The models process data through separate spatial (image-based) and temporal (sequence-based) streams. These models illustrate how multi-stream deep learning architectures can effectively analyze the temporal morphological changes that occur during embryo development. Abbasi et al. \cite{abbasi2023time} employed a novel approach for video data classification by transforming video data into multivariate time series. Their method emphasizes the morphological changes of the fetus over time and connects these changes to the outcomes. To enhance prediction accuracy, they modified time series classifiers by incorporating attention mechanisms that can capture both short-term and long-term dependencies. The proposed method demonstrates promising results in prediction.
Sharma et al. \cite{sharma2024deep} proposed an AI system designed to analyze past embryo development through 2-hour video sequences and predict future morphological changes for up to 23 hours. This is achieved using a LSTM-based predictive model, which recursively forecasts future video frames by utilizing temporal patterns in embryo dynamics. This approach allows for an early assessment of the embryo's developmental potential. 
Typically, embryo transfer occurs in clinics on day 5, during the blastocyst stage. However, performing the transfer earlier, on day 3, can increase the chances of a successful pregnancy. In this context, Kalyani et al. \cite{kalyani2024deep} introduce a novel ResNet-GRU deep learning model designed to predict blastocyst formation at 72 HPI. They utilize ResNet for extracting spatial features and GRU for analyzing temporal patterns in time-lapse images. This research aids embryologists in identifying the most suitable embryos for transfer on day 3, thereby improving patient outcomes and increasing pregnancy rates in assisted reproductive technology (ART).

 \begin{figure*}[t]
    \centering
    \includegraphics[width=\textwidth, clip, trim={0 7 0 0}]{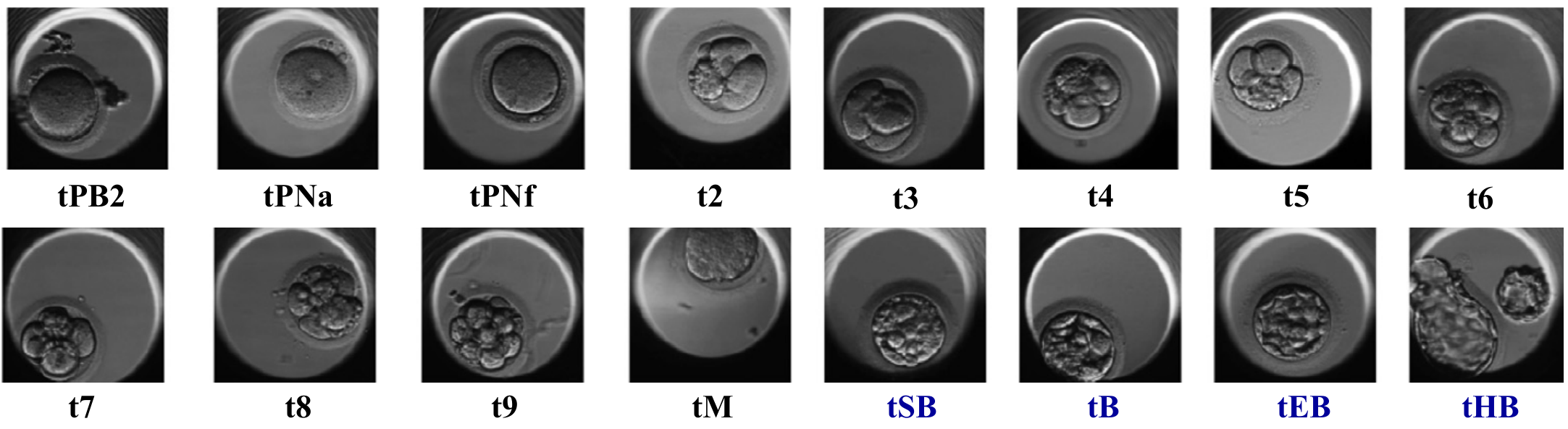}
    \caption{Time-lapse images of the 16 stages of embryonic development \cite{gomez2022time}.} 
    \label{fig:Data} 
\end{figure*}

Mohamed et al. \cite{mohamed2023automated} proposed an automated system for blastocyst embryo quality grading that uses transfer learning and VGG-16 with novel classification layers to enhance model efficiency. In the first stage, preprocessing and data augmentation are performed, and classification and evaluation are performed in the second stage. 
Mazroa et al. \citep{mazroa2024anomaly} proposed the Embryo Development and Morphology using a Computer Vision-Aided Swin Transformer with Boosted Dipper-Throated Optimization (EDMCV-STBDTO) technique for accurate detection of embryo development. The process begins with image preprocessing using a bidirectional filter (BF) model to eliminate noise. This is followed by feature extraction using the Swin Transformer. Additionally, a variable autoencoder method is applied for data classification. The effectiveness of the EDMCV-STBDTO method is validated through comprehensive studies utilizing benchmark datasets.
Kim et al. \cite{kim2024multimodal} introduced a multimodal deep learning model that combines time-lapse video data with Electronic Health Records (EHRs) to predict embryo viability. This approach reduces the subjectivity and time demands associated with conventional manual embryo assessment in clinical IVF.

Xie et al. \cite{xie2022early} introduced the Attentive Multifocus Selection Network (AMSNet) to predict blastocyst formation from early time-lapse images. The model integrates multi-focus images to preserve in-depth information and uses a feature channel shift mechanism to capture long-term temporal dependencies. Garg et al. \cite{garg2024efficient} developed a model that combines the strengths of InceptionV3 and DenseNet201. They used data augmentation to address the issue of class imbalance. Next, they utilized the InceptionV3 and DenseNet201 models in parallel, incorporating additional layers such as global average pooling, dense layers with ReLU activation, and dropout layers to enhance the model's performance. Asghari Varzaneh et al. \cite{varzaneh2025lightweight} proposed a model that integrates DINOv2 for spatial feature extraction with an efficient video vision Transformer for temporal analysis of sparse embryo image sequences. By reducing dependence on full time-lapse monitoring, this study offers a practical solution to support embryologists in making more informed embryo selection decisions.

Despite recent advancements, accurately predicting blastocyst formation from sparse daily images, which is often a practical limitation in many IVF labs, continues to be a significant challenge. This study aims to address this issue by proposing a novel DINOv2-LSTM architecture with multi-head attention, specifically designed to achieve high accuracy using only these limited daily images.

\section{Dataset Description}\label{sec:Dataset Description}
\begin{table}
\small
\centering
\caption{\small Definitions of development stage labels assigned to individual images}
\label{tab:Dev}
\begin{tabular}{l l @{\hspace{5em}} l l} 
\hline
\textbf{Stage} & \textbf{Definition} & \textbf{Stage} & \textbf{Definition} \\
\hline
tPB2 & second polar body appearance & t7 & seventh cell appears \\
tPNa & pronuclei appearance & t8 & eight cell appears \\
tPNf & pronuclei fading & t9 & ninth cell appears \\
t2 & second cell appears & tM & morula formation \\
t3 & third cell appears & tSB & blastulation start \\
t4 & fourth cell appears & tB & blastocyst formation \\
t5 & fifth cell appears & tEB & blastocyst expansion \\
t6 & sixth cell appears & tHB & blastocyst hatching \\
\hline
\end{tabular}
\end{table}

The dataset utilized in this study comprises time-lapse embryo recordings from 716 infertile couples who underwent intracytoplasmic sperm injection (ICSI) cycles at an IVF center \cite{gomez2022time}. To facilitate continuous monitoring of embryo development, all embryos were cultured in time-lapse imaging incubators(TLI), with images captured by a camera every 10 to 20 minutes and The resolution of an embryoscope image is 500$\times$500 pixels. The dataset, collected by Tristan et al., contains 704 carefully annotated video samples. Out of these, 499 videos correspond to embryos deemed suitable for transfer, while the remaining samples were discarded due to poor quality or insufficient growth. Each video contains 16 consecutive stages of embryonic development, annotated by a qualified embryologist. A definition of each developmental stage can be found in Table ~\ref{tab:Dev}. The annotations cover the cell division stages up to 9 cells, followed by the stages of compaction and blastulation.  Also, Figure \ref{fig:Data} depicts the 16 stages of embryonic development.

To label videos, we assign a positive label (Blastocyst) to sequences that are at least in one of the stages of blastocyst formation or growth, including tSB, tB, tEB, and tHB, and otherwise a negative label (non-Blastocyst). There are 522 video samples in the Blastocyst class, while the non-Blastocyst class contains 182 samples.

\section{Blastocyst Formation}
\begin{figure*}[t]
  \centering
  \includegraphics[width=\textwidth]{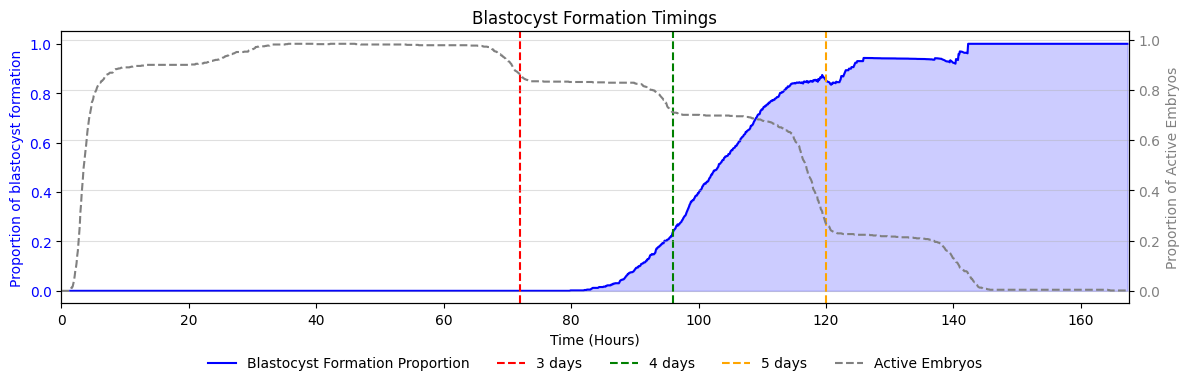}
  \caption{\small Proportion of blastocyst formation over time. The blue line tracks the proportion of blastocyst-formatted embryos out of the total 704 embryos. The grey-dashed line follows the Active Embryos over time (Notice that some of the embryos are not annotated until +24 h).}
  \label{fig:BF} 
\end{figure*}

Successfully reaching the blastocyst stage of an embryo is a critical milestone in IVF development that directly impacts clinical outcomes \cite{practice2018blastocyst}. As shown in Figure \ref{fig:BF}, the proportion of embryos in the dataset reaching blastocyst formation follows a distinct temporal pattern that is closely related to morphological competence. Before day 3 (marked by the red line in Figure \ref{fig:BF}), blastocyst formation is biologically impossible, as embryos remain in the cleavage stage. However, the transition beyond this point initiates a fundamental change.

After day 3, the proportion of blastocyst formation increases rapidly, reaching a peak between days 5 and 7. During this time period, embryos that show stable growth have a significantly higher probability of achieving blastocyst morphology. 

The long culture period acts as a biological filter: embryos that stop growing (due to chromosomal abnormalities, metabolic stress, or adverse conditions) do not progress, while those that are intrinsically growth competent reach blastocysts. As a result, viable blastocysts detected by day 5 to 7 are strongly associated with improved implantation potential and live birth rates. However, the probability of blastocyst formation is not simply time-dependent, and factors such as embryo quality, laboratory conditions, patient genetic factors, etc., influence blastulation potential. Although the period after day 3 is crucial for blastocyst development, the outcome depends on a complex mix of biological, technical, and clinical factors.

\section{Material and methods}
\begin{figure}[t]
    \centering
    \includegraphics[width=\textwidth]{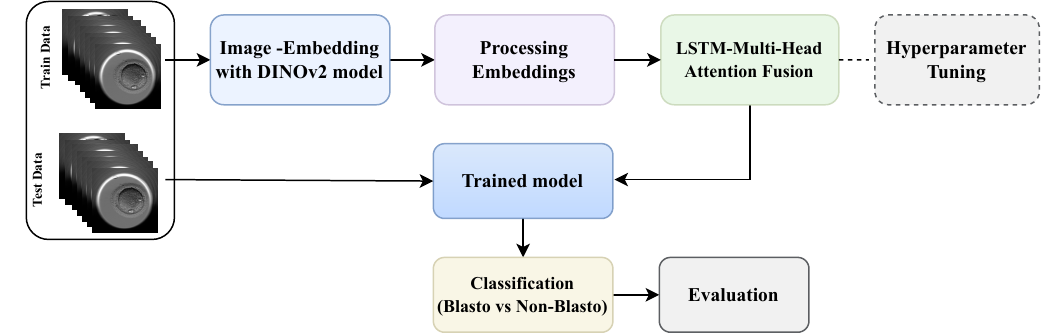}  
    \caption{\small An overview of our proposed hybrid model:The process begins with DINOv2 extracting features from embryo frames, followed by temporal analysis using an LSTM with Multi-Head Attention and hyperparameter tuning. The model then classifies sequences into Blasto or Non-Blasto categories.} 
    \label{fig:example} 
\end{figure}

In this paper, we propose a hybrid time-series model for embryo development prediction. Our framework starts with segmenting and preprocessing frames, followed by feature extraction using DINOv2, where embeddings are adjusted to better match the specific characteristics of the data and the problem under study. To improve temporal modeling, we integrate a multi-head attention mechanism into an LSTM backbone to enhance long-range dependencies across sequential frames. Additionally, hyperparameter tuning is performed to optimize the model's performance. Figure \ref{fig:example} shows an overview of the proposed model, and the details of each component are explained in the following sections.

\subsection{Preprocessing data frames}
Each data sample is a time-lapse video that shows how the embryo grows during 5 to 6 days. These videos are continuously recorded to track how the embryo grows during this time.
To make our analysis model, we take at most $m = 7$ frames from each video. We call this set $F = \{f_1, f_2, ..., f_m\}$. The first frame $f_1$ shows the start of embryo growth. The last frame $f_m$ shows the end of growth.
The other frames in the set are selected based on a fixed time interval. We use the (Eq.~\ref{eq: seg}) as follows:
\begin{equation} 
F_i = F_1 + (i - 1) \times \Delta t 
    \label{eq: seg}
\end{equation}
where $\Delta t = 24$ hours and $i = 2, ..., 6$. This means that each frame is chosen every 24 hours. This method ensures that we cover the full development period evenly, and it also follows clinical routines that use 24-hour intervals between observations.

After choosing the frames, we start the preprocessing step to prepare the images for feature extraction. First, each raw embryo image is converted into a grayscale image. This makes important structures in the image clearer and also reduces the amount of data that needs to be processed.
Next, all images are resized to $518 \times 518$ pixels. This resizing step is necessary to make the images compatible with the input requirements of the DINOv2 feature extractor.

\subsection{Image embedding with DINOv2}

DINOv2 \cite{oquab2023dinov2} is a self‑supervised ViT‑based foundation model that learns image representations without any manual labeling. These general‑purpose visual features extracted from its ViT backbone can be used for tasks such as classification, segmentation, and depth estimation with minimal fine‑tuning. DINOv2 is pre‑trained on LVD‑142M, a curated dataset of 142 million unlabeled images filtered and deduplicated from an initial 1.2 billion web images, ensuring high quality and diversity.

The training of DINOv2 involves a teacher–student approach, where the teacher’s weights are updated as an exponential moving average (EMA) of the student’s weights, ensuring the teacher provides consistent training signals without direct backpropagation \cite{oquab2023dinov2}.

DINOv2 integrates two complementary learning objectives, each operating with its own prediction head attached to both the student and teacher ViT backbone:

\begin{itemize}
  \item \textbf{Image‑level objective:}  
    The student’s \texttt{[CLS]} embedding on a local crop is trained to match the teacher’s \texttt{[CLS]} embedding on a global crop, fostering invariance to transformations and high‑level semantic understanding \cite{oquab2023dinov2}.

  \item \textbf{Patch‑level objective:}  
    A random subset of input patches is masked in the student stream, and the student is trained to predict the teacher’s patch embeddings at those positions, encouraging fine‑grained, locality‑aware features crucial for dense prediction tasks \cite{oquab2023dinov2}.
\end{itemize}

By iteratively optimizing these objectives over the vast unlabeled dataset, DINOv2 learns a rich, hierarchical visual representation. Once pre‑trained, the ViT backbone provides (i) a global \texttt{[CLS]} token embedding and (ii) a sequence of per‑patch embeddings for each image. In our pipeline, we extract both the \texttt{[CLS]} vector and the average‑pooled per‑patch embeddings, concatenate them into a single descriptor to creat a \(1\times1536\) embedding vector and use this vector as our final image representation for all downstream tasks.

\begin{figure}[t]
    \centering
    \includegraphics[width=0.85\textwidth, height=9cm]{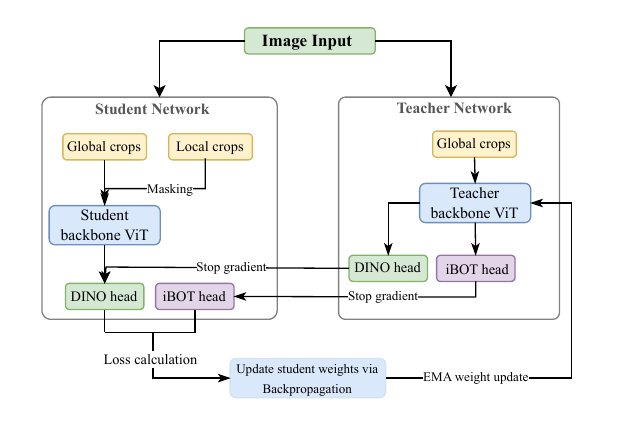}  
    \caption{\small A framework of DINOv2 architecture} 
    \label{fig:example2} 
\end{figure}

\subsection{Post-processing embedding vectors}
The embedding vectors extracted from the DINOv2 model cannot be used directly in our temporal classification system. Before passing into the model, two essential preprocessing steps are required: normalize the feature values and make sure all the sequences have the same length. This is important because our classification models expect input data with a fixed shape.

In real situations, the number of frames or time steps $L$ in each sequence is different. Some sequences are short, and some are long, because some embryos may not develop fully, and their growth may stop in the first few days. To handle this inconsistency, we use a method called zero-padding. This means that we fill empty places with zeros at the end of shorter sequences, so that all sequences have the same length. We call this target length $L_{\text{max}}$, which is the length of the longest sequence. This method follows the approach from \cite{hashemi2019enlarging}.

We utilize Eq.~\ref{eq: Embd2} to employ the padding process. Once this step has been done, we get a final data structure written as $\mathbf{X} \in \mathbb{R}^{N \times L_{\text{max}} \times 1536}$. Here, $N$ is the number of sequences we have in a batch, $L_{\text{max}}$ is the length we set for all sequences after padding, and 1536 is the size of each feature vector that comes from DINOv2 for every single frame in the sequence.

\begin{equation}
    \mathbf{E}_{\text{padded}} =
    \begin{cases}
    [\mathbf{e}_1, ..., \mathbf{e}_L, \mathbf{0}, ..., \mathbf{0}] & \text{if } L < L_{\text{max}} \\
    [\mathbf{e}_1, ..., \mathbf{e}_{L_{\text{max}}}] & \text{if } L \geq L_{\text{max}}
    \end{cases}
    \label{eq: Embd2}
\end{equation}

In this equation, $\mathbf{E}_{\text{padded}}$ is the new sequence after padding or trimming. Each vector $\mathbf{e}_t \in \mathbb{R}^{1536}$ is the embedding from DINOv2 at time $t$, and it contains important information about the embryo at that stage. The value $t=1$ shows the first time step, and $t=L$ shows the last one in the original sequence. This way, our model can work with all sequences in the same way, even if they had different lengths at the beginning. The full pipeline for extracting and processing features with DINOv2 is illustrated in Figure~\ref{fig:example2}.

\subsection{LSTM-Multi-Head Attention Fusion for Temporal Sequence Modeling}
We propose an enhanced sequential modelling architecture that integrates LSTM layers with a Multi-Head Attention (MHA) mechanism to improve the temporal modelling and contextual understanding of sequential frame-level features. The model is designed to classify visual embedding sequences extracted from pre-trained DINOv2. The model accepts as input a sequence $X \in \mathbb{R}^{T \times D}$, where $T$ is the number of frames in a sequence and $D$ is the feature dimension of each frame (in our case, $D=1536$). The architecture proceeds through stacked LSTM layers, followed by a multi-head attention module and a classification head. 

In the context of embryonic development classification, LSTM layers enable the model to capture dynamic temporal patterns, such as gradual morphological changes over time, which are vital for accurate stage prediction \cite{yu2019review}. Let $X = \{x_1, x_2, \dots, x_T\}$ denote a sequence of frame-based visual embeddings, where each $x_t \in \mathbb{R}^D$ is extracted from DINOv2 as a high-dimensional pretrained visual transformer. The LSTM unit updates its hidden state $h_t$ and cell state $c_t$ at each time step $t$ \cite{neil2016phased}. This mechanism enables the model to maintain and update the temporal memory of embryonic structures, such as shape changes and cell divisions, that gradually evolve over time. By modeling such patterns in sequence, LSTMs improve the predictive power of the developmental stage.

Although LSTMs capture temporal continuity well, they are limited in modeling non-local dependencies over time. For example, anomalies in the early frame may be correlated with the final stage results, which LSTMs may not exploit well \cite{al2024rnn}. To address this issue, we integrate a MHA mechanism \cite{cordonnier2020multi} on top of the LSTM outputs. MHA allows the model to re-evaluate the contribution of each frame with respect to all other frames in the sequence.

\begin{figure}[t]
    \centering
    \includegraphics[width=\textwidth, clip,trim={0 30 0 280}]{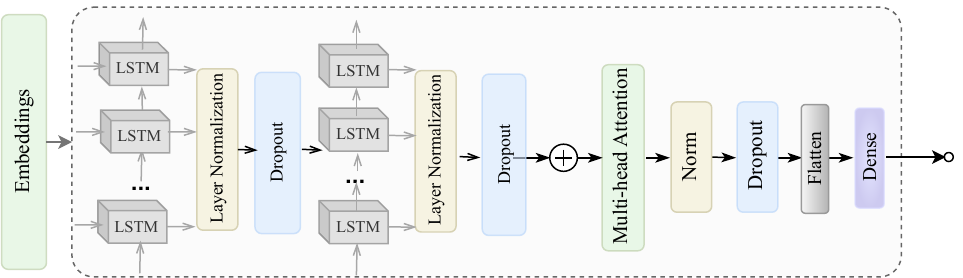}  
    \caption{\small A framework of LSTM-Multi-Head Attention fusion. An architecture combining stacked LSTMs for temporal feature extraction with Multi-Head Attention to capture long-range dependencies, followed by normalization and classification layers.} 
    \label{fig:example3} 
\end{figure}

Given the output of the LSTM layers $H = [h_1, h_2, \dots, h_T] \in \mathbb{R}^{T \times d}$, we project it into queries $Q$, keys $K$, and values $V$ using learned linear transformations as calculated in (Eq.~\ref{eq: Att}):
\begin{equation} 
Q = HW^Q, \quad K = HW^K, \quad V = HW^V
\label{eq: Att}
\end{equation}
Each attention head computes the scaled dot-product attention independently in (Eq.~\ref{eq: Att2}) as follows:
\begin{equation} 
\text{Attention}(Q, K, V) = \text{softmax}\left(\frac{QK^\top}{\sqrt{d_k}}\right)V
\label{eq: Att2}
\end{equation}

where $d_k$ is the dimensionality of the key vectors. In the multi-head setting, $h$ parallel attention heads are used to allow the model to jointly attend to information from different representation subspaces:
\begin{equation} 
\text{MHA}(H) = \text{Concat}(\text{head}_1, \dots, \text{head}_h)W^O
\end{equation}

where each $\text{head}_i = \text{Attention}(Q_i, K_i, V_i)$ with separate learned projections $W_i^Q$, $W_i^K$, $W_i^V$ for the $i$-th head, and $W^O$ is the output projection matrix.
Using $h_t$ as the source and target of attention enables the model to consider contextual signals from distant but semantically related time frames \cite{lipton2015learning}.
This is particularly valuable in the analysis of embryonic development, especially when distinctive features such as the appearance of the blastocoel or changes in cell density may appear transiently and not necessarily in consecutive frames.
The output of the attention module is combined with the LSTM representation via a residual connection and normalized to stabilize training. Subsequently, the temporally aggregated features are flattened and passed through a classification head. This head maps the learned temporal context to a binary decision (Blastocyst vs. non-Blastocyst). Overall, integrating sequential modeling with LSTMs and dynamic reweighting through MHA, the architecture effectively models both local continuity and long-range dependencies, making it well-suited for the complex task of embryonic stage classification.\\
To address class imbalance between Blastocyst ($y_i=1$) and non-Blastocyst ($y_i=0$) samples, we employ a weighted variant binary cross-entropy (Eq.~\ref{eq:weighted_loss}) with class weights $w_{y_i} $ to balance gradient contributions during training.
\begin{equation}
\begin{split}
\mathcal{L}_{\text{weighted}} = & -\frac{1}{N}\sum_{i=1}^N w_{y_i}\left[y_i \log(p_i) \right. \\
& \left. + (1-y_i)\log(1-p_i)\right]
\end{split}
\label{eq:weighted_loss}
\end{equation}
where $N$ is the total number of samples and $p_i$ is the predicted probability of sample $i$ belonging to class 1. A framework of the proposed LSTM-Multi-Head Attention is presented in Figure \ref{fig:example3}.

\section{Experimental Results}
To address our research questions—RQ1 on hybrid architecture performance and RQ2 on the contribution of the multi-head attention mechanism, we conducted experiments using the dataset described in Section~\ref{sec:Dataset Description}. First, preprocessing was applied to the time-lapse embryo videos. From each video, a maximum of 7 frames were selected. Each selected frame was converted to grayscale and resized to meet the input requirements of the DINOv2 feature extractor. Then, we selected the best hyperparameters for each model. After that, we evaluated the models using different performance measures. In the next sections, we present the details of the model implementation and the evaluation results, and discuss comparisons with other experiments.

\subsection{Experimental setup}
\textbf{Hyperparameter setting:}
To keep the input image sizes the same, we resized all of them to $518 \times 518 $ before training and testing the model. The training was done with a learning rate of 0.001, using a batch size of 16, and continued for 20 epochs. To reduce the risk of overfitting, we added dropout with a rate of 0.3. Also, if the chosen validation metric does not improve, the training stops after 20 steps of no change. The dataset was randomly split into two parts: 80\% for training and 20\% for testing.

\textbf{Evaluation metrics:} To evaluate the model's performance, we used common evaluation metrics such as accuracy, precision, recall, and F1-score \cite{naidu2023review}. Besides that, we also created a confusion matrix and a ROC curve to show how well the model performs in classification \cite{vujovic2021classification}.

\subsection{Results and Discussion}
\begin{figure}[t]
\centering
\begin{minipage}{0.48\textwidth}
\centering
\includegraphics[width=\textwidth, height=5.5cm]{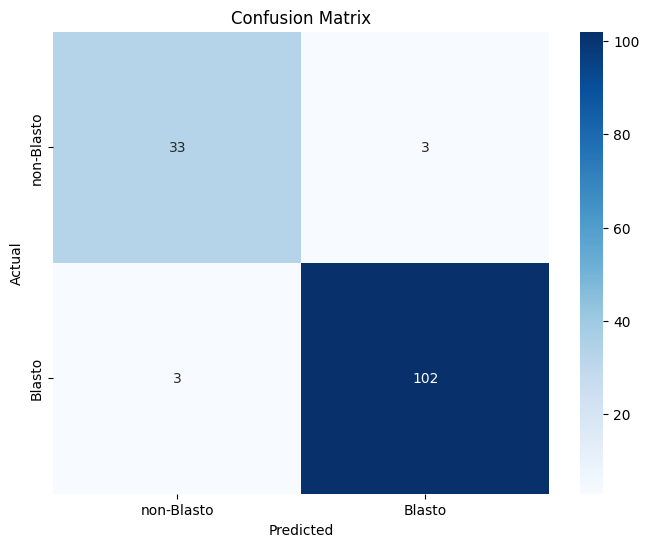}
\caption{Confusion matrix for predicting classes}
\label{fig:confusion}
\end{minipage}
\hfill
\begin{minipage}{0.48\textwidth}
\centering
\includegraphics[width=\textwidth, height=5.5cm]{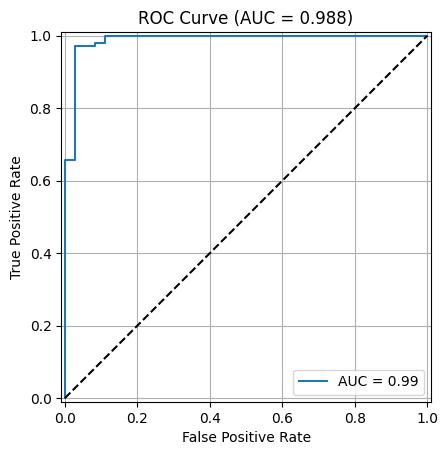}
\caption{ROC curve of LSTM-Multi-Head Attention Fusion}
\label{fig:roc}
\end{minipage}
\end{figure}

To classify data into blastocyst and non-blastocyst categories, we developed a sequence-based model. This model predicts the class of an embryo by analyzing the sequence of frames captured during its development. The results of the proposed model are detailed below.\\
\textbf{Experiments with LSTM-MHA fusion:} The evaluation results of this model on the selected prediction frames indicate strong discriminatory power between the two classes in the test images. The precision, recall, and F1-score metrics were calculated separately for both classes, yielding values of 0.971 for the blastocyst class and 0.918 for the non-blastocyst class, with the blastocyst class slightly outperforming the non-blastocyst class. These results were obtained from the average of 10 independent runs. Figure \ref{fig:confusion} illustrates the model's performance in classifying the two classes by displaying the confusion matrix.

\begin{figure}[t]
    \centering
  \includegraphics[width=0.9\textwidth]{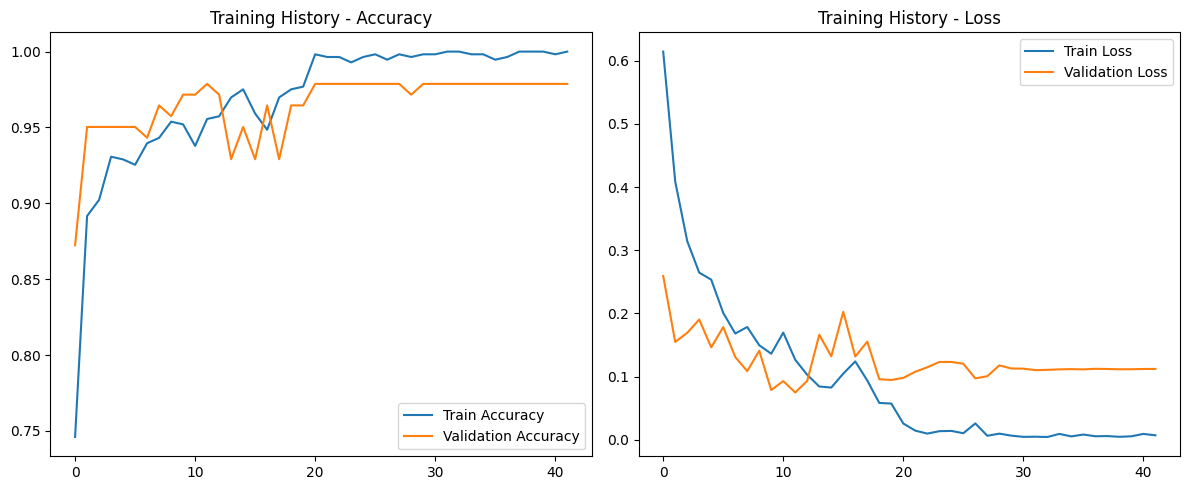} 
    \caption{The training history for loss and accuracy metrics. An early stopping patience of 30 epochs was set; however, as the curves show, the model converged and showed no significant improvement after about epoch 20, confirming the early termination of training to prevent overfitting.} 
    \label{fig:History} 
\end{figure}

An overall accuracy of 0.964 further confirms the model's reliability in classification tasks. The training history for both loss and accuracy metrics is presented in Figure \ref{fig:History}. Additionally, Figure \ref{fig:roc} displays the ROC Curve, with the AUC (Area Under the ROC Curve) reaching 0.99. This suggests that the model possesses exceptional discriminative ability between the two classes, exhibiting a very low rate of false positives and false negatives.\\
In recent years, many researchers have used AI, machine learning, and deep learning methods to analyze time-lapse imaging of embryo development. However, similar analyses on our specific data type are scarce; existing studies are typically limited to either blastocyst images or clinical patient data. Kalyani et al. \cite{kalyani2024deep} have worked on a similar data set to ours, using all image frames from embryo development up to day 3 to predict whether or not a blastocyst will form.
Tables~\ref{tab:metrics} and~\ref{tab:accuracy} summarize the classification performance of different models used for blastocyst classification. Table ~\ref{tab:metrics} reports the calculated precision, recall, and F1 score by class for the proposed LSTM-MHA Fusion, as well as the model proposed by Kalyani et al., while Table~\ref{tab:accuracy} presents the overall accuracy and the type of classification for each model. 
The performance metrics of the proposed model are reported as mean ± standard deviation across 10 independent runs. The low standard deviation values indicate consistent and stable performance across all runs, demonstrating the reliability of the proposed model.

The comparison of models indicates that the proposed LSTM-MHA Fusion architectures achieve a high overall accuracy 96.4\%. However, a deeper inspection of class-wise metrics reveals that the LSTM-MHA model provides a better balance between precision and recall across both blastocyst and non-blastocyst classes, which is critical for minimizing false positives and false negatives in medical predictions.

\begin{table}
\small
\centering
\caption{Performance comparison of different models for each class}
\label{tab:metrics}
\renewcommand{\arraystretch}{1.1}
\begin{tabular}{@{}llccc@{}}
\toprule
\textbf{Model} & \textbf{Class} & \textbf{Precision} & \textbf{Recall} & \textbf{F1-Score} \\
\midrule
\multirow{2}{*}{LSTM-MHA Fusion} & Blastocyst & $0.971 \pm 0.016$  & $0.971 \pm 0.018$ & $0.971 \pm 0.014$ \\
                                 & Non-Blastocyst & $0.918 \pm 0.015$ & $0.918 \pm 0.015$ & $0.918 \pm 0.009$ \\
\midrule
\multirow{2}{*}{Kalyani et al.\cite{kalyani2024deep}} & Blastocyst & 0.93 & 0.98 & 0.95 \\
                                & Non-Blastocyst & 0.91 & 0.77 & 0.83 \\
\bottomrule
\end{tabular}
\end{table}

\begin{table}
\centering
\caption{Accuracy and classification types of different models}
\label{tab:accuracy}
\renewcommand{\arraystretch}{1.1}
\begin{tabular}{@{}lcc@{}}
\toprule
\textbf{Model} & \textbf{Classification Type} & \textbf{Accuracy} \\
\midrule
CNN & Binary & 0.64 \\
CNN with GA & Binary / Multiclass & 0.72 / 0.35 \\
DenseNet201--LSTM & Binary & 0.76 \\
ResNet50--LSTM (Kalyani et al.\cite{kalyani2024deep}) & Binary & 0.85\\
ResNet50--GRU (Kalyani et al.\cite{kalyani2024deep}) & Binary & 0.93 \\
\textbf{LSTM-MHA Fusion} & Binary & \textbf{ $0.964 \pm 0.017$ }\\
\bottomrule
\end{tabular}
\end{table}

Although the ResNet50--GRU model from Kalyani et al. achieves a high accuracy of 0.93, its recall for the non-blastocyst class (0.77) is significantly lower, indicating potential weaknesses in detecting negative cases. The CNN and DenseNet-based models lag significantly behind in accuracy and are less suitable for reliable classification. In conclusion, the LSTM-MHA Fusion model has the best trade-off between overall accuracy and class-wise performance. As a result,  it is the most robust choice among the evaluated approaches.

\textbf{Experiments on other benchmark datasets:} 
Due to the unavailability of other datasets specifically related to our study's domain, the proposed model was evaluated on several alternative time-series datasets to assess its generalizability. A brief description of each dataset is provided, followed by a detailed presentation and analysis of the experimental results. \\
\textbf{Car:} This dataset contains 1-D time series representations of vehicle outlines (sedans, pickups, minivans, and SUVs), extracted from traffic video streams using motion analysis. The core objective is to perform classification of these unique temporal signatures into their respective vehicle categories \cite{kaggle_Car}.\\
\textbf{Financial Distress:} This dataset comprises panel data for a sample of companies, tracking each across 1 to 14 time periods. It contains 83 lagged features (x1-x83), including financial metrics and one categorical variable, to classify companies as financially distressed (1) or healthy (0) based on a defined threshold of the "Financial Distress" target variable \cite{kaggle_Financial}.\\
\textbf{Occupancy Detection:} Time-series sensor data (temperature, humidity, light, CO2) is used to classify room occupancy. Ground-truth labels were accurately generated from minute-by-minute pictures \cite{UCI_Occupancy}.\\
\textbf{EEG Eye State:} This dataset contains a 117-second continuous EEG recording from a single session, featuring 14 sensor values measured by an Emotiv Neuroheadset. The corresponding eye state (1 for closed, 0 for open) was manually annotated frame-by-frame from synchronized camera footage and serves as the classification target \cite{UCI_EEG}.

The comprehensive results across different evaluation metrics are presented in Table~\ref{tab:4}.
As shown in the table, the model demonstrated exceptional performance on the Occupancy detection dataset. It achieved a classification accuracy of 0.99. Furthermore, the precision, recall, and F1-score for both classes are excellent, approaching the ideal value of 1.00.
The results on the Financial analysis data are also relatively strong, indicating the model's robustness in a different application context.
Experiments conducted on the EEG data yielded slightly lower, yet still acceptable, results compared to the other datasets. The detailed metrics for all classes can be observed in the table.
Evaluation on the Car dataset, which comprises four classes, resulted in an overall classification accuracy of 0.72. The performance varies across classes; for instance, the second class achieved an F1-score of approximately 0.92, which is significantly higher than the F1-score for other classes. This discrepancy is likely attributable to the common issue of class imbalance within the dataset, which often leads to such variations in per-class performance.

 In addressing the research questions presented in Section \ref{sec:introduction}, we evaluated the key aspects that express the efficiency and effectiveness of our approach. The high overall accuracy of 0.964 and an AUC of 0.99 provide a direct response to RQ1, \textit{confirming that the hybrid DINOv2 and LSTM architecture is effective in predicting blastocyst formation from daily images}. Additionally, in response to RQ2, \textit{we observed that the multi-head attention mechanism significantly enhances the LSTM's ability to focus on critical developmental stages}. This improvement results in a more robust predictor compared to standard sequential models.
In summary, the experimental results on these diverse time-series datasets demonstrate that the proposed model possesses a significant degree of generalizability and is capable of performing effectively across various data types.
\begin{table}
\centering
\small
\caption{Performance evaluation of the proposed model on various time-series datasets.}
\label{tab:4}
\begin{tabular}{lcccccc}
\hline
\textbf{Dataset} & \textbf{Class} & \textbf{Precision} & \textbf{Recall} & \textbf{F1-Score} & \textbf{Accuracy} \\
\hline
\multirow{2}{*}{Financial Distress} & Class 0 & 0.96 & 0.92 & 0.94 & \multirow{2}{*}{0.92} \\
 & Class 1 & 0.84 & 0.93 & 0.88 & \\
\hline
\multirow{2}{*}{Occupancy Detection} & Class 0 & 1.00 & 0.99 & 1.00 & \multirow{2}{*}{0.99} \\
 & Class 1 & 0.98 & 1.00 & 0.99 & \\
\hline
\multirow{2}{*}{EEG Eye State} & Class 0 & 0.82 & 0.84 & 0.83 & \multirow{2}{*}{0.81} \\
 & Class 1 & 0.79 & 0.78 & 0.79 & \\
\hline
\multirow{4}{*}{Car} & Class 0 & 0.57 & 0.57 & 0.57 & \multirow{4}{*}{0.72} \\
 & Class 1 & 1.00 & 0.86 & 0.92 & \\
 & Class 2 & 0.78 & 0.74 & 0.76 & \\
 & Class 3 & 0.56 & 0.69 & 0.62 & \\
\hline
\end{tabular}
\end{table}

\subsection{Ablation study}
In this section, we evaluate the impact of incorporating a multi-head attention mechanism into the basic LSTM model. The standard model classifies frames into two categories: \textit{Blasto} and \textit{non-Blasto}. To enhance the model's performance, we introduced a multi-head attention layer on top of the LSTM outputs. This addition allows the model to focus more effectively on the dependencies and relationships between frames. The performance of the models is compared using several evaluation metrics, including precision, recall, the F1 score for each class, and overall accuracy.
\begin{table}
\centering
\small
\caption{Comparison of LSTM and LSTM-MHA Fusion Models}
\label{tab:ablation}
\begin{tabular}{lcccc}
\toprule
\textbf{Model} & \textbf{Class} & \textbf{Precision} & \textbf{Recall} & \textbf{F1-Score} \\
\midrule
\multirow{2}{*}{LSTM} 
  & Blasto     & 0.944 & 0.968 & 0.956 \\
  & non-Blasto & 0.923 & 0.869 & 0.895 \\
\midrule
\multirow{2}{*}{LSTM-MHA Fusion} 
  & Blasto     & 0.971 & 0.971 & 0.971 \\
  & non-Blasto & 0.918 & 0.918 & 0.918 \\
\midrule
\multicolumn{5}{l}{\footnotesize \textbf{Overall Accuracy:} \hspace{0.5cm}  LSTM: 93.86\%  , \hspace{0.5cm}  LSTM-MHA Fusion: 96.43\%} \\
\bottomrule
\end{tabular}
\end{table}
As shown in Table \ref{tab:ablation}, the LSTM-MHA fusion model outperforms the base LSTM model across all metrics. The overall accuracy of the base model is 93.86\%, while the accuracy of the LSTM-MHA fusion model is 96.43\%, representing an increase of 2.57\%. For the \textit{Blasto} class, the F1-score of the proposed model is 0.971, compared to 0.956 for the base model. These improvements are also significant for the \textit{non-Blasto} class.

\section{Limitations}
Although this study provides a detailed analysis of sequential images of embryo development in an IVF setting, there is a significant limitation that impacts the generalizability of the findings. The primary limitation arises from the severe lack of publicly available, high-quality datasets documenting human embryo development using TLI across multiple IVF laboratories. Our analysis relied exclusively on the \cite{gomez2022time} data, which is the only substantial, real-world TLI dataset currently available to the research community. This exclusive reliance severely limits the generalizability of our results to analyses of data generated across multiple laboratories. Ultimately, this lack of shared, multicenter TLI data is not only a limitation of this specific study, but also a broader challenge for the field, potentially slowing progress toward standardized, universally applicable models for embryo selection and understanding developmental changes. To address this limitation, a collaborative effort will be needed in the future to enhance multi-laboratory partnerships and establish shared, ethical data repositories.

\section{Conclusion}
In this study, we proposed a novel technique for classifying embryo development into blastocyst and non-blastocyst formation. The method involves compressing the sequence of embryonic frames into a shorter segment, followed by classification using sequence-based models. Among the approaches tested on the examined data in this study, the proposed LSTM-MHA Fusion demonstrated superior performance compared to existing methods. The findings suggest that applying such models can significantly reduce human error and support the selection of the most viable embryos for transfer. Future research could extend this work by assessing the quality of the formed blastocysts based on inner cell mass (ICM) and trophectoderm (TE) features. Furthermore, to ensure the reliability and generalizability of the proposed models, they should be validated on diverse real-world datasets. Incorporating additional factors such as patient clinical profiles and environmental conditions in IVF laboratories could also enhance the robustness of embryo assessment systems.

\section*{Acknowledgements}
\noindent{\footnotesize This study is conducted as part of the EIVF-AI project funded by Vinnova, the Swedish Governmental Agency for Innovation Systems (Grant No.2024-01462).}

\vspace{0.5em}
\section*{Author Contributions}
\noindent{\footnotesize \textbf{Zahra Asghari Varzaneh} and \textbf{Niclas Wölner-Hanssen:} Conceptualization, Data gathering, Methodology, Formal analysis, Software, Validation, Visualization, Investigation, Writing—original draft. \textbf{Reza Khoshkangini:} Conceptualization, Formal analysis, Investigation, Methodology, Project administration, Supervision, Writing—review and editing. \textbf{Thomas Ebner:} Conceptualization, Formal analysis, Investigation, Methodology, Supervision, Writing—review and editing.
\textbf {Magnus Johnsson:} Conceptualization, Formal analysis, Investigation, Methodology, Project administration, Supervision, Writing—review and editing.
All authors read and approved the final manuscript.}

\vspace{0.5em}
\section*{Data Availability}
\noindent{\footnotesize The datasets analysed during the current study are available in the Human embryo time-lapse video dataset repository, https://doi.org/10.5281/zenodo.7912264}






\bibliographystyle{elsarticle-num} 
\bibliography{references}





\end{document}